\pdfoutput=1

\documentclass[11pt]{article}

\usepackage{emnlp2021}

\usepackage{times}
\usepackage{latexsym}

\usepackage[T1]{fontenc}

\usepackage[utf8]{inputenc}

\usepackage{microtype}

\usepackage{xcolor}
\usepackage{soul}
\usepackage{tabularx}
\usepackage{graphicx}
\usepackage{multirow}
\usepackage{amsmath}
\usepackage{amsfonts}
\usepackage[british, english]{babel}
\usepackage{CJKutf8}
\usepackage{CJK}
\usepackage{enumitem}
\usepackage{float}

\usepackage{booktabs}
\title{Data Augmentation with Hierarchical SQL-to-Question Generation for Cross-domain Text-to-SQL Parsing}

\author{
Kun Wu$^1$\footnotemark[1],
Lijie Wang$^2$,
Zhenghua Li$^1$,
Ao Zhang$^2$,\\
\textbf{Xinyan Xiao}$^2$,
\textbf{Hua Wu}$^2$,
\textbf{Min Zhang}$^1$,
\textbf{Haifeng Wang}$^2$\\
1. Institute of Artificial Intelligence, School of Computer Science and Technology,\\
Soochow University, Suzhou, China \\
2. Baidu Inc, Beijing, China \\
kwu@stu.suda.edu.cn,
\{zhli13,minzhang\}@suda.edu.cn \\
\{wanglijie,zhangao,xiaoxinyan,wu\_hua,wanghaifeng\}@baidu.com\
}

\begin{document}
\maketitle

\renewcommand{\thefootnote}{\fnsymbol{footnote}}
\footnotetext[1]{Work done during an internship at Baidu Inc.}
\renewcommand{\thefootnote}{\arabic{footnote}}

\begin{abstract}
Data augmentation has attracted a lot of research attention in the deep learning era for its ability in alleviating data sparseness. The lack of labeled data for unseen evaluation databases is exactly the major challenge for cross-domain text-to-SQL parsing. Previous works either require human intervention to guarantee the quality of generated data, or fail to handle complex SQL queries. This paper presents a simple yet effective data augmentation framework. First, given a database, we automatically produce a large number of SQL queries based on an abstract syntax tree grammar. For better distribution matching, we require that at least 80\% of SQL patterns in the training data are covered by generated queries. Second, we propose a hierarchical SQL-to-question generation model to obtain high-quality natural language questions, which is the major contribution of this work. Finally, we design a simple sampling strategy that can greatly improve training efficiency given large amounts of generated data. Experiments on three cross-domain datasets, i.e., WikiSQL and Spider in English, and DuSQL in Chinese, show that our proposed data augmentation framework can consistently improve performance over strong baselines, and the hierarchical generation component is the key for the improvement.






\end{abstract}

\section{Introduction}
\begin{figure}[tb]
\centering
\includegraphics[width=0.43\textwidth]{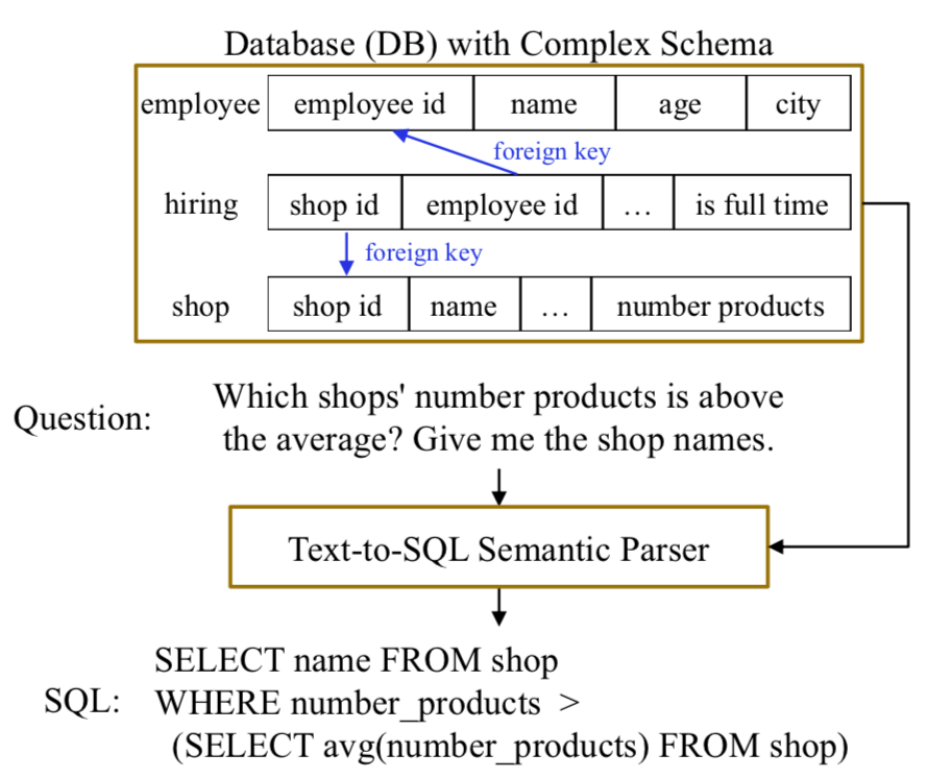}
\caption{An example of the text-to-SQL parsing task.}
\label{fig:intro}
\end{figure}

Given a natural language (NL) question and a relational database (DB), the text-to-SQL parsing task aims to produce a legal and executable SQL query to get the correct answer~\cite{c1997guide}, as depicted in Figure~\ref{fig:intro}. A DB usually consists of multiple tables interconnected via foreign keys. 

Early research on text-to-SQL parsing mainly focuses on the in-domain setting~\cite{li2014constructing,iyer2017learning,yaghmazadeh2017sqlizer}, where all question/SQL pairs of train/dev/test sets are generated against the same DB. In order to deal with the more realistic setting where DBs in the evaluation phase are unseen in the training data, researchers propose several cross-domain datasets, such as  WikiSQL~\cite{zhong2017seq2sql} and Spider~\cite{yu2018spider} in English, and DuSQL~\cite{wang-etal-2020-dusql} in Chinese. All three datasets adopt the DB-level data splitting, meaning that a DB and all its corresponding question/SQL pairs can appear in only one of the train/dev/test sets. 

\begin{figure*}[tb]
\centering
\includegraphics[width=0.98\textwidth]{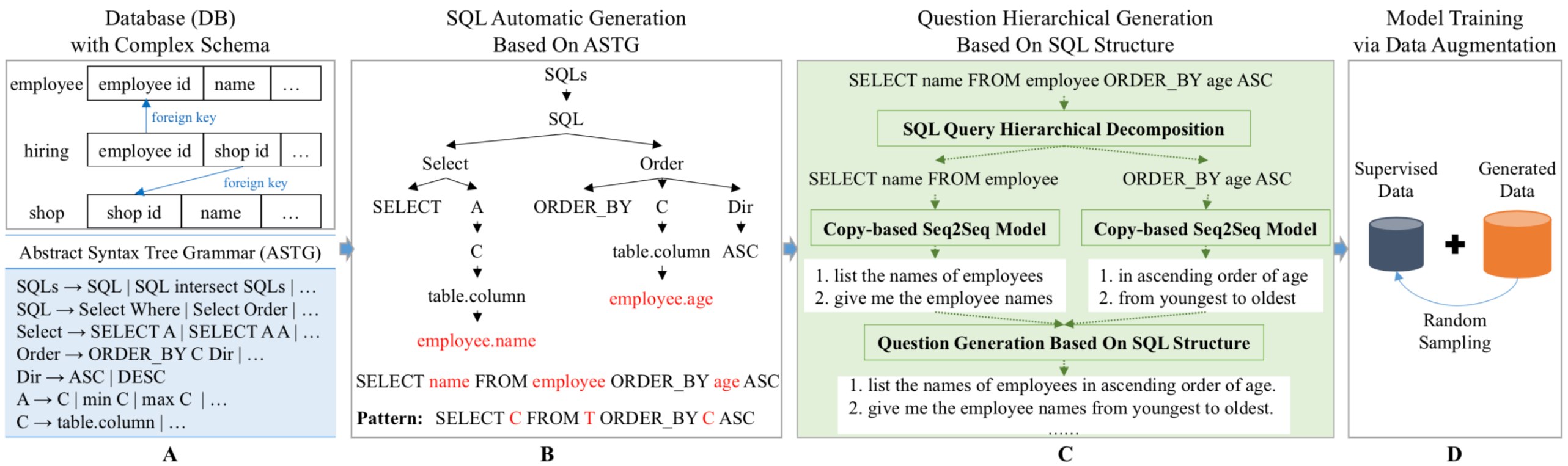}
\caption{An overview of our approach containing 3 stages: SQL query generation based on ASTG (Section $\S$\ref{ssec:sql_generation}), question hierarchical generation according to SQL structure (Section $\S$\ref{ssec:ques_gen}), model training via data augmentation (Section $\S$\ref{ssec:train_aug}).}
\label{fig:intro_case}
\end{figure*}

Cross-domain text-to-SQL parsing has two major challenges. First, unseen DBs usually introduce new schemas, such as new table/column names and unknown semantics of inter-table relationships. Therefore, it is crucial for a parsing model to have strong generalization ability. The second challenge is that the scale of labeled data is quite small for such a complex task, since it is extremely difficult to construct DBs and manually annotate corresponding question/SQL pairs. For example, the Spider dataset has only 200 DBs and 10K question/SQL pairs in total.

To deal with the first challenge, many previous works focus on how to better encode the matching among questions and DB schemas, and achieve promising performance gains~\cite{sun2018semantic,guo2019towards,rat-sql}. 

To handle the second challenge of lacking in labeled data, and inspired by the success of vanilla pretraining models with masked language model loss~\cite{devlin2019bert}, researchers propose task-specific pretraining models for semantic parsing~\cite{yin-etal-2020-tabert,herzig-etal-2020-tapas,yu2020grappa,shi2020learning}. The basic idea is to learn joint representations of structured data (i.e., tables) and corresponding contextual texts by designing delicate objective losses with large amounts of collected data that is related to the target task. 
These customized models have achieved good performance on English datasets. However, pretraining is slow and expensive as the models are trained on millions of web tables and related contexts. In addition, these approaches are currently only experimented on English since it is difficult to collect such data for pretraining.

This work follows another research line, i.e., data augmentation, which addresses both challenges discussed above in a resource-cheap way. The idea of data augmentation is automatically generating noisy labeled data using some deliberately designed method, and the technique has been successfully applied to a wide range of NLP  tasks~\cite{barzilay2001extracting,jia-liang-2016-data}. In our cross-domain text-to-SQL task, we can directly generate labeled data over unseen DBs as extra training data. The key of data augmentation is how to improve the quality of generated data. As two prior works, 
\citeauthor{yu2018syntaxsqlnet} \shortcite{yu2018syntaxsqlnet} manually align question tokens and DB elements in the corresponding SQL query, in order to obtain relatively high-quality question/SQL pairs,
while \citeauthor{guo2018question} \shortcite{guo2018question} utilize a flat Seq2Seq model to directly translate SQL queries to NL questions, which may only work for simple queries (see Section $\S$\ref{sec:related-work} for detailed discussion). 

This work proposes a data augmentation framework with hierarchical SQL-to-question generation in order to obtain higher-quality question/SQL pairs. The framework consists of two steps. First, given a DB, we use an abstract syntax tree grammar (ASTG) to automatically generate SQL queries. For better distribution matching, we require the generated queries to cover at least 80\% of SQL patterns in the original training data. Second, we design a hierarchical SQL-to-question generation model to obtain NL questions. The basic idea is: 1) decomposing a SQL query into clauses according to its syntax tree structure; 2) translating each clause into a subquestion; 3) concatenating subquestions into a full question according to the execution order of the SQL query.
Finally, we design a simple sampling strategy to improve training efficiency with augmented data.
In summary, we make the following contributions.  
\begin{itemize}[leftmargin=*]
\item We present a simple and resource-cheap data augmentation framework for cross-domain text-to-SQL parsing with no human intervention.\footnote{We release the code at \url{https://github.com/PaddlePaddle/Research/tree/master/NLP/Text2SQL-DA-HIER}.}  
\item As the key component for our framework, we propose a hierarchical SQL-to-question generation model to obtain more reliable NL questions.
\item In order to improve training efficiency, we propose a simple sampling strategy to utilize generated data, which is of relatively larger scale than original training data.
\item We conduct experiments and analysis on three datasets in both English and Chinese, i.e., WikiSQL, Spider, and DuSQL, showing that our proposed framework can consistently improve performance over strong baselines. 
\end{itemize}

\section{Proposed Data Augmentation Approach}
\label{sec:method}
Given a DB, the goal of data augmentation is to automatically generate high-quality question/SQL pairs as extra training data. The key for its success lies in two aspects. First, the generated SQL queries should have similar distribution with the original data. Second, generated NL questions reflect the meaning of the corresponding SQL queries, especially for complex queries.

Our proposed framework adopts a two-step generation process, as shown in Figure~\ref{fig:intro_case}. We first generate SQL queries at different complexity levels based on an ASTG, and then translate SQL queries into NL questions using our proposed hierarchical generation model.

\begin{figure}[tb]
\centering
\includegraphics[width=0.48\textwidth]{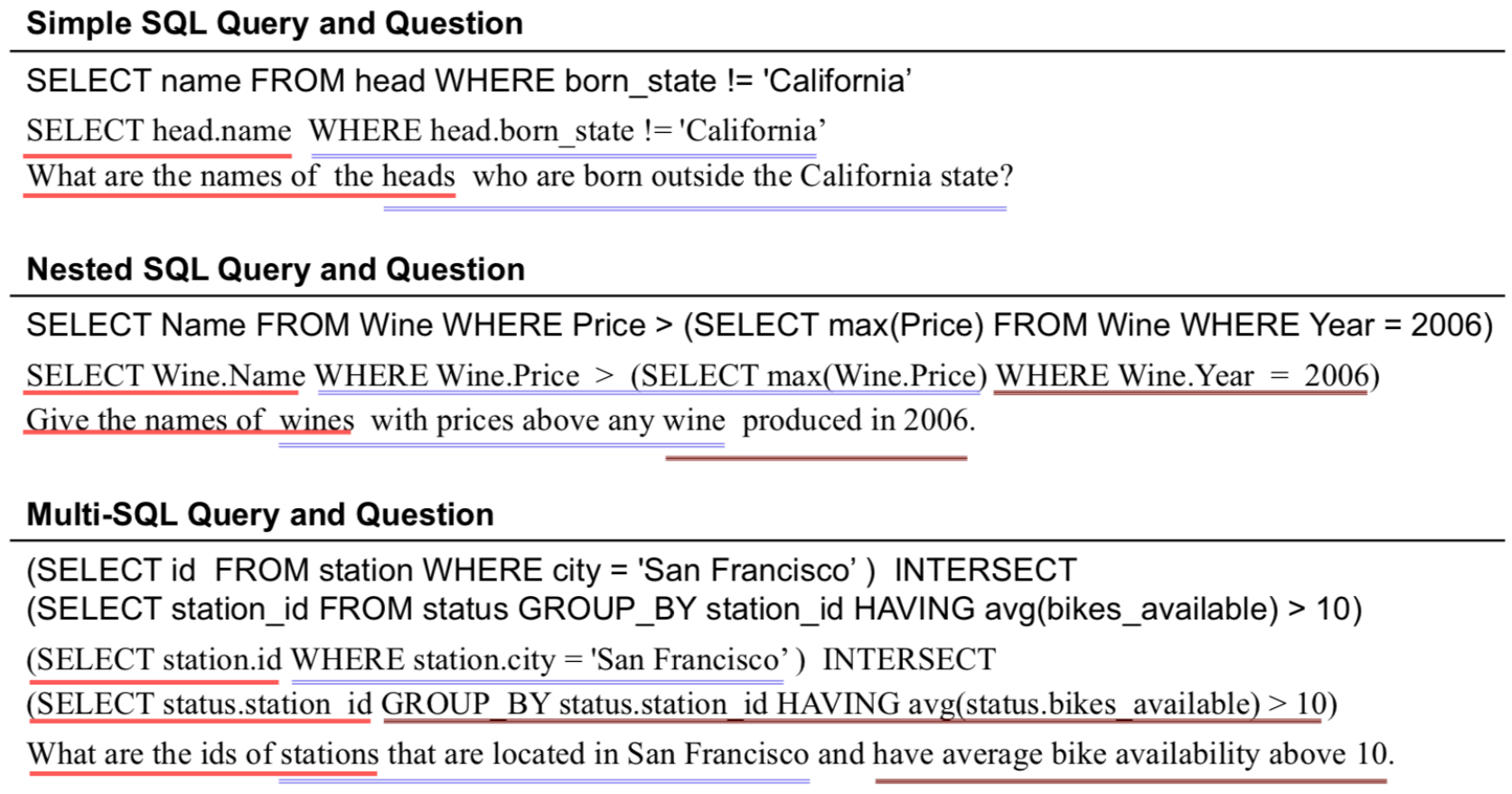}
\caption{Examples of segment-level mapping between SQL queries and corresponding questions from the Spider dataset. In each example, the second SQL query is the equivalent of the first.}
\label{fig:align_exa}
\end{figure}

\subsection{SQL Query Generation}
\label{ssec:sql_generation}
Being a program language, all SQL queries can be represented as nested tree structures, as depicted in Figure~\ref{fig:intro_case}-B according to some context-free grammar. In fact, most text-to-SQL parsers proposed recently adopt the abstract syntax tree representation at the decoding stage~\cite{yin2018tranx,yu2018syntaxsqlnet,guo2019towards,rat-sql}. Following those works, we design a general ASTG that can cover all SQL patterns in our adopted benchmark datasets. Due to space limitation, Figure~\ref{fig:intro_case} shows a fraction of the production rules. 

According to our ASTG, the SQL query in Figure~\ref{fig:intro_case}-B can be generated using the production rules: ``\textit{SQLs $\rightarrow$ SQL}'', ``\textit{SQL $\rightarrow$ Select Order}'', ``\textit{Select $\rightarrow$ SELECT A}'', ``\textit{Order $\rightarrow$ ORDER\_BY C Dir}'', etc.

By assembling production rules from our ASTG, we can generate any \emph{sketch tree}. As shown in Figure~\ref{fig:intro_case}-B, a sketch tree means that DB-related leaf nodes (marked in red) are removed, and its flat form corresponds to \emph{a pattern}, shown at the bottom. In our work, we generate sketch trees from simple to complex. Under a certain complexity level, i.e., tree breadth and depth, we first generate all possible sketch trees, and then apply them to new DBs to produce full trees (i.e., SQL queries) by filling DB-related items, such as table names, column names, and cell values.

In order to better match the query distribution of real text-to-SQL training data and to limit the number of generated SQL queries as well, we stop sketch tree generation when the generated ones cover more than 80\% of patterns in the original training data\footnote{As discussed in the logic form-based semantic parsing work of \citeauthor{herzig2019don} \shortcite{herzig2019don}, distribution mismatch is mainly caused by insufficient coverage of logical form templates.}. 
The SQL queries are generated in a way that simpler SQL patterns come first, and 80\% of the remaining patterns are usually high-frequency patterns. This limitation aims to control the complexity of generated questions, since very complex questions are rare in the training data. 
Please kindly note that our simple ASTG-based generation procedure can produce a lot of patterns unseen in the original data, because our generation is at production rule level. This is advantageous from the data variety perspective.  

Moreover, given a DB, we only keep executable SQL queries for correctness check. 

\subsection{Hierarchical SQL-to-Question Generation}
\label{ssec:ques_gen}

Given an SQL query, especially a complex one, it is difficult to generate an NL question that represents exactly same meaning. In their data augmentation work, \citeauthor{guo2018question} \shortcite{guo2018question} use a vanilla Seq2Seq model to translate SQL queries into NL questions and obtain performance boost on WikiSQL consisting of simple queries. However, as shown in Table~\ref{tab:all_results}, we find performance consistently drops on all datasets over our strong baselines, which is largely due to the quality issue of generated NL questions, as illustrated in Table \ref{tab:case study}.

This work proposes a hierarchical SQL-to-question generation model to produce higher-quality NL questions. The idea is motivated by our observation that there is a strong segment-level mapping between SQL queries and corresponding questions, as shown in Figure~\ref{fig:align_exa}. For example, the SQL query of the first example can be decomposed into two segments, i.e., the SELECT clause and the WHERE clause. The two clauses naturally correspond to the two question segments, i.e., ``What are the names of the heads'' and ``heads who are born outside the California state'' respectively. 

Following the observation, our hierarchical SQL-to-question generation consists of three steps: 1) decomposing the given SQL query into several clauses; 2) translating every clause into a subquestion; 3) combining subquestions into a full NL question. Next we describe each step in detail. 

\textbf{Step 1: SQL clause decomposition}. We decompose an SQL query into multiple clauses based on SQL keywords. 
Usually a clause contains only one keyword. 
In some cases multiple keywords are put into the same clause according to semantics. More formally speaking, multiple keywords are combined into one single clause based on two perspectives of consideration: 1) SQL syntax, and 2) alignment between SQL queries and NL questions, as illustrated by the last two examples in Figure \ref{fig:align_exa}.

From the perspective of SQL syntax, HAVING and GROUP$\_$BY, are naturally bundled together, and thus are put into one clause, as shown in the third example of Figure \ref{fig:align_exa}. LIMIT and ORDER$\_$BY are similarly handled.

From the second perspective, some keywords are not explicitly expressed in NL questions. In other words, there is a mismatch between intents expressed in NL questions and the implementation details in SQL queries. To better align them, we follow IRNet \cite{guo2019towards} and combine GROUP$\_$BY with either SELECT or ORDER$\_$BY.

For a nested SQL query, e.g., the second example in Figure \ref{fig:align_exa}, it is more reasonable to put the outside WHERE and the inside SELECT into one clause, since they together express a complete operation semantically. 


Based on our decomposition method, an unseen SQL pattern always consists of common clause patterns in the training data.

\textbf{Step 2: clause-to-subquestion translation}. Compared with a full SQL query, a clause has a flat structure and involves simple semantics corresponding to a single SQL operation. Thus, it is much easier to translate clauses to subquestions compared with direct SQL-to-question translation. We use a standard copy-based Seq2Seq model~\cite{gu2016incorporating} for clause-to-subquestion generation. The details are presented in Section $\S$\ref{ssec:train_detail}.

\textbf{Step 3: question composition}. As shown in Figure~\ref{fig:align_exa}, we compose a full question by concatenating all subquestions in a certain order. We experiment with two ordering strategies, i.e., the execution order of corresponding clauses\footnote{The execution order used in our work is: WHERE -> GROUP BY -> HAVING -> SELECT, ORDER\_BY -> LIMIT. For the SQL query with a nested part, we firstly generate the question for the nested part and use it as a fragment of the whole question, also known as the bottom-up way.}, and the sequential order of corresponding clauses in the full SQL query. Preliminary experiments show that the former performs slightly better, which is thus adopted in our framework. Please note that direct concatenation may lead to redundant words from adjacent subquestions. We use several heuristic rules to handle this. Taking the third multi-SQL query in Figure \ref{fig:align_exa} for example, since the two SELECT clauses are translated into nearly the same subquestion, we only keep one in the final NL question. 

\textbf{Discussion on quality of generated NL questions.} Our reviewers suggest to evaluate naturalness and truth of generated NL questions. Due to time limitation, we did not perform strict manual evaluation. So far, our approach mainly considers the informativeness aspect of generated NL questions. We leave such evaluation and analysis as future work, which will certainly help us better understand our proposed approach.

\subsection{Clause-to-subquestion Translation Model}
\label{ssec:train_detail}
We adopt the standard Seq2Seq model with copy mechanism \cite{gu2016incorporating} for clause-to-subquestion translation, which is also used in our baseline, i.e., flat SQL-to-question translation, with the same hyper-parameter settings. 

In the input layer, we represent every SQL token by concatenating two embeddings, i.e., word (token as string) embedding, and token type (column/table/value/others) embedding, each having a dimension size of 150. 
We use default values for other hyper-parameters.

\textbf{Training data construction}. We construct clause/subquestion pairs for training the translation model from the original training data, consisting of two steps. The first step decomposes a SQL query into clauses using the same way illustrated above.

The second step aims to decompose an NL question into clause-corresponding subquestions. In other words, this step finds a subquestion (i.e., a segment of the question) for each clause. We first build alignments between tokens in the SQL query and the corresponding NL question based on simple string matching. 
The string-matching method\footnote{We extract all question n-grams ($1 \le n \le 6$) to match DB elements (i.e., columns, tables, and values) in the corresponding SQL query, so as to get alignments between question n-grams and DB elements.} is very similar to the schema linking step in IRNet \cite{guo2019towards} and RATSQL \cite{rat-sql}.
Then for each clause, we define the corresponding subquestion as the shortest question segment that contains all DB elements in the clause.
Finally, we discard low-confidence clause/subquestion pairs to reduce noises, such as subquestions having large overlap with others. 
We keep overlapping subquestions, unless one subquestion fully contains another. In that case, we only keep the shorter subquestion.

We find that a portion of collected clauses  have multiple subquestion translations. For example, the clause ``\textit{ORDER\_BY age ASC}'' are translated as both ``in ascending order of the age'' and ``from youngest to oldest''. We follow \citeauthor{hou2018sequence}~\shortcite{hou2018sequence} and use them as two independent clause/subquestion pairs for training. 

\subsection{Three Strategies for Utilizing Generated Data}
\label{ssec:train_aug}

Given a set of DBs, the generated question/SQL pairs are usually of larger scale than the original training data (see Table~\ref{tab:num_data}), which may greatly increase training time. 
In this work, we compare the following three strategies for parser training. 
\begin{itemize}[leftmargin=*]
\item \textbf{The pre-training strategy} first pre-trains the model with only generated data, and then fine-tunes the model with labeled training data.
\item \textbf{The directly merging strategy} trains the model with all generated data and labeled training data in each epoch. 
\item \textbf{The sampling strategy} first randomly samples a number of generated data and trains the model on both sampled and labeled data in each epoch. The sampling size is set to be the same with the size of the labeled training data. 
\end{itemize}

\begin{table}[tb]
\small
\center
\begin{tabularx}{0.40\textwidth}{l|r|r}
\hline Dataset & Labeled Data & Generated Data \\
\hline
WikiSQL & 61,297 & 98,206\\
Spider  & 8,625 & 58,691\\
DuSQL & 18,602 & 45,942\\
\hline
\end{tabularx}
\caption{Data statistics in the number of question/SQL pairs. The column of labeled data shows the number of training pairs.}
\label{tab:num_data}
\end{table}

\section{Experiments}

\paragraph{Datasets.} We adopt three widely-used cross-domain text-to-SQL parsing datasets to evaluate the effectiveness of different approaches, i.e., \emph{WikiSQL}\footnote{\url{https://github.com/salesforce/WikiSQL}} and \emph{Spider}\footnote{\url{https://yale-lily.github.io/spider}} in English, and \emph{DuSQL}\footnote{\url{https://aistudio.baidu.com/aistudio/competition/detail/47}} in Chinese. All datasets follow their original data splitting. WikiSQL focuses on single-table DBs and simple SQL queries that contain only one SELECT clause with one WHERE clause. In contrast, Spider and DuSQL are much more difficult in the sense each DB contains many tables and SQL queries may contain advanced operations such as clustering, sorting, calculation (only for DuSQL) and have nested or multi-SQL structures.

For each dataset, we generate a large number of question/SQL pairs against the evaluation DBs. Table~\ref{tab:num_data} shows the size of the generated data. It is noteworthy that since the Spider test data is not publicly released, we generate data and evaluate different approaches against the Spider-dev DBs and question/SQL pairs. 

\paragraph{Baseline parsers.} We choose four popular open-source parsers to verify our proposed frameowrk.

\textbf{WikiSQL: SQLova.} The SQLova parser \cite{hwang2019comprehensive} achieves competitive performance on WikiSQL without using execution guidance and extra knowledge (e.g., DB content and other datasets). The encoder obtains table-aware representations by applying BERT to concatenated sequence of question and table schema, and the decoder generates SQL queries as slot filling in the SELECT/WHERE clauses. HydraNet \cite{lyu2020hybrid} reported the state-of-the-art (SOTA) performance on WikiSQL but did not release their code. 

\textbf{Spider: IRNet and RATSQL}. IRNet \cite{guo2019towards} is an efficient yet highly competitive parser for handling complex SQL queries on Spider, consisting of two novel components: 1) linking DB schemas with questions via string matching; 2) a grammar-based decoder to generate SemQL trees as intermediate representations of SQLs. RATSQL \cite{rat-sql} is the current SOTA parser on Spider. The key contribution is utilizing a relation-aware transformer encoder to better model the connections between DB schemas and NL questions. However, training RATSQL is very expensive. It takes about 7 days to train a basic BERT-enhanced RATSQL model on a V100 GPU card, which is about 10 times slower than IRNet. 

With limited computational resource, we mainly use IRNet for model ablation and efficiency comparison. Meanwhile, we report main results on RATSQL to learn the effect of our proposed framework on more powerful parsers. 

Another detail about RATSQL to be noticed is that we use the released Version 2 (V2). They reported higher performance with V3\footnote{The comparison of V2 and V3 is discussed at \url{https://github.com/microsoft/rat-sql/issues/12}.} by better hyper-parameter settings and even longer training time. However, they did not release their configurations.

\textbf{DuSQL: IRNet-Ext}. IRNet-Ext proposed by \citet{wang-etal-2020-dusql} is an extended version of IRNet to accommodate the characteristics of Chinese dataset DuSQL. In this work, we further enhance IRNet-Ext with BERT. Basically, we concatenate the NL question and DB schema as the input, and perform encoding with BERT (instead of BiLSTM in the original parser). 

\paragraph{Evaluation metrics.} 
We use the exact matching (EM) accuracy as the main metric, meaning the percentage of questions whose predicted SQL query is equivalent to the gold SQL query, regardless of clause and component ordering. 
We also use component matching (CM) F1 score to evaluate the clause-level performance for in-depth analysis.
Besides, we report execution (exec) accuracy on WikiSQL, meaning the percentage of questions whose predicted SQL query obtains the correct answer. 

\paragraph{Hyper-parameter settings.} 
For each parser, we use default parameter settings in their released code. 
All these parsers are enhanced with vanilla (in contrast to task-specific) pretraining models, i.e., BERT~\cite{devlin2019bert}, including IRNet-Ext.

In order to avoid the effect of {\color{black}{performance vibrations}}\footnote{Please see issues proposed at the github of RATSQL model, such as \url{https://github.com/microsoft/rat-sql/issues/10}.}, we run each model for 5 times with different random initialization seeds, and report the averaged EM accuracy (mean)
and the variance (
$\sqrt{\frac{\sum_{i=1}^n(x_i-\bar{x})^2}{n-1}}$). We only run each RATSQL model for 3 times due to its prohibitively high requirement on computational resource. 

\defcitealias{guo2019towards}{Guo2019}
\defcitealias{lyu2020hybrid}{Lyu2020}
\defcitealias{hwang2019comprehensive}{Hwang2019}
\defcitealias{guo2018question}{Guo2018}
\defcitealias{rat-sql}{Wang2020a}
\defcitealias{yu2018syntaxsqlnet}{Yu2018a}
\defcitealias{wang-etal-2020-dusql}{Wang2020b}

\begin{table}[tb]
\small
\center
\begin{tabularx}{0.46\textwidth}{l|l}
\toprule[2pt]
\multicolumn{2}{c}{\textbf{WikiSQL}} \\
\cline{1-2}
Models & EM [Exec] \\
\hline
HydraNet (\citetalias{lyu2020hybrid}) & 83.8 [89.2] \\
SQLova (\citetalias{hwang2019comprehensive}) & 80.7 [86.2] \\
STAMP (\citetalias{guo2018question}) & 60.7 [74.4] \\
\qquad $+$ Aug (FLAT) & 63.7 ($+3.0$) [75.5] \\
\hline
SQLova (ours) & 80.1$_{\pm 0.40}$ [85.7] \\
\qquad $+$ Aug (FLAT) & 79.7$_{\pm 0.50}$ ($-0.4$) [85.4] \\
\qquad $+$ Aug (HIER) & 81.2$_{\pm 0.09}$ ($+1.1$) [86.5] \\
\midrule[1pt]
\multicolumn{2}{c}{\textbf{Spider}}\\
\cline{1-2}
Models & EM\\
\hline
IRNet (\citetalias{guo2019towards}) & 60.6 \\
RATSQL V3 (\citetalias{rat-sql}) & 69.6 \\
RATSQL V2 (\citetalias{rat-sql}) & 65.8 \\
SyntaxSQLNet (\citetalias{yu2018syntaxsqlnet}) & 22.1 \\
\qquad $+$ Aug (PATTERN) & 28.7 ($+6.6$) \\
\hline
IRNet (ours) & 59.7$_{\pm 0.41}$ \\
\qquad $+$ Aug (FLAT) & 58.8$_{\pm 0.56}$ ($-0.9$)\\
\qquad $+$ Aug (HIER) & 61.8$_{\pm 0.32}$ ($+2.1$)\\
RATSQL V2 (ours) & 65.4$_{\pm 0.60}$\\
\qquad $+$ Aug (HIER) & 68.2$_{\pm 0.42}$ ($+2.8$)\\
\midrule[1pt]
\multicolumn{2}{c}{\textbf{DuSQL}}\\
\cline{1-2}
Models & EM\\
\hline
IRNet-Ext (\citetalias{wang-etal-2020-dusql}) & 50.1 \\
\hline
IRNet-Ext + BERT (ours) & 53.7$_{\pm 0.60}$ \\
\qquad $+$ Aug (FLAT) & 53.4$_{\pm 0.67}$ ($-0.3$)\\
\qquad $+$ Aug (HIER) & 60.5$_{\pm 0.40}$ ($+6.8$) \\
\bottomrule[2pt]
\end{tabularx}
\caption{Main results. We run each model for 5 times, and report the average and variance (as subscripts). $+$Aug means the model is enhanced with data augmentation, and \textit{PATTERN}, \textit{FLAT}, and \textit{HIER} refer to the three data augmentation approaches.}
\label{tab:all_results}
\end{table}

\begin{table}[tb]
\footnotesize
\center
\begin{tabularx}{0.43\textwidth}{l|l}
\hline 
SQL & SELECT draw\_size FROM matches \\ 
 & WHERE loser\_age $>$ 10 \\
FLAT & what are the percentage of draw size in matches \\
 & with loser higher than 10? \\
HIER & with losers who are older than 10, \\
 & find the draw size of the matches.\\
\hline
SQL & SELECT horsepower FROM cars\_data \\
 & WHERE edispl $<=$ 10 \\
 & ORDER\_BY year DESC \\
FLAT & list all horsepower year \\
 & in descending order of year. \\
HIER & with edispl no higher than 10, \\
 & show the horsepower of the cars, \\
 & made from most recently to oldest.\\
\hline
\end{tabularx}
\caption{Case study on FLAT vs. HIER on a simple SQL query (first) and a complex one (second) from Spider.}
\label{tab:case study}
\end{table}

\begin{table*}[tb]
\renewcommand\tabcolsep{2.5pt}
\centering
\begin{small}
\begin{tabular}{l | l | lllll}
\toprule
Datasets & Models & SELECT & WHERE & GROUP\_BY & HAVING & ORDER\_BY\\
\hline
\multirow{2} {*} {WikiSQL} & SQLova & 88.1 & 90.2 & -- & -- & -- \\
 &\qquad $+$ Aug & 87.8 ($-0.3$)& 91.6 ($+1.4$) & -- & -- & -- \\
\hline
\multirow{4} {*} {Spider} & IRNet & 87.7 & 68.0 & 80.8 & 75.5 & 76.3 \\
 &\qquad $+$ Aug & 88.5 ($+0.8$) & 70.1 ($+2.1$)& 81.0 ($+0.2$)& 77.2 ($+1.7$)& 80.0 ($+4.5$) \\
& RATSQL & 85.5 & 72.6 & 79.3 & 76.7 & 79.4 \\
 &\qquad $+$ Aug & 87.7 ($+2.2$) & 75.4 ($+2.8$)& 82.5 ($+3.2$)& 79.4 ($+2.7$)& 80.9 ($+1.5$) \\
\hline
\multirow{2} {*} {DuSQL} & IRNet-Ext & 78.5 & 82.4 & 93.8 & 92.1 & 93.4\\
 &\qquad $+$ Aug & 79.8 ($+1.3$) & 86.6 ($+4.2$) & 95.0 ($+1.2$)& 93.3 ($+1.1$) & 93.5 ($+0.1$)\\
\bottomrule
\end{tabular}
\end{small}
\caption{CM F1 scores over five types of SQL clauses. The type division is borrowed from \citeauthor{yu2018spider}\shortcite{yu2018spider}.}
\label{tab:complex_exp}
\end{table*} 

\subsection{Main Results}
\label{ssec:main_result}
Table~\ref{tab:all_results} shows the main results. For each dataset, the first major row shows previously reported results, and the second major row gives results of our base parsers without and with data augmentation. 

To compare previous data augmentation methods, we also re-implement the flat one-stage generation approach (FLAT) proposed by \citet{guo2018question}. We do not implement the pattern-based data augmentation approach (PATTERN) of \citet{yu2018syntaxsqlnet} due to its requirement of human intervention. Moreover, their large performance improvement is obtained over a very weak baseline. 

\textbf{Performance of our baseline parsers.} On WikiSQL, the averaged performance of our SQLova parser is lower than their reported performance by about 0.7. On Spider, the performance of our IRNet parser is lower than their reported value by 0.9. However, please kindly note that we use default configurations of SQLova and IRNet, and our best results among five runs on both WikiSQL and Spider are very close to theirs. 

As discussed earlier, HydraNet and RATSQL v3 achieve higher performance, but they do not release their code or configurations. 

In summary, we can conclude that our baseline parsers achieve competitive results on all three datasets. We believe that it would be more reasonable to report the mean and variance of performance. 

\textbf{Comparison of different data augmentation methods.} According to our results in the second major row of each dataset, data augmentation with FLAT leads to consistent performance degradation, which is contradictory to the results on WikiSQL reported by \citet{guo2018question}. We suspect the reason is that our BERT-enhanced baseline parser is much stronger than their adopted parser. To verify this, we run SQLova without BERT and find similar performance gains from 61.0\% to 64.0\% via the FLAT data augmentation. Using HIER, the performance can further increase to 66.1\%. Due to time and resource limitation, we do not run similar experiments on the other two datasets. 

In contrast to FLAT, our proposed HIER approach achieves consistent improvement over the strong BERT-enhanced parsers. In particular, it is very interesting to see that the parsers have lower performance variance compared with the baselines. We will give more insights on the effectiveness of the hierarchical generation approach in Section $\S$\ref{ssec:analysis}. Again, to save computational resource, we did not run RATSQL with the FLAT data augmentation approach. 

Looking closer into the improvements on the three datasets, we can see that our HIER data augmentation obtains the least performance increase on WikiSQL, possibly due to the higher baseline performance with relatively large-scale labeled data consisting of simple SQL queries. The most gain is obtained on DuSQL. We suspect the reason is two-fold. First, the baseline performance is the lowest, which is similar to the results obtained by \citet{yu2018syntaxsqlnet} on Spider with data augmentation. Second, during the construction of DuSQL, \citet{wang-etal-2020-dusql} first automatically generate question/SQL pairs and then perform manual correction and paraphrasing, leading to certain resemblance between their labeled data and our generated data. 

In summary, we can conclude that our proposed augmentation approach with hierarchical SQL-to-question generation is more effective than previous methods, and can substantially improve performance over strong baselines, especially over complex datasets. In the future, we would like to apply our approach to other text-to-SQL datasets and languages.

\begin{table}[tb]
\small
\center
\begin{tabularx}{0.42\textwidth}{l|l|l}
\hline Models & Seen patterns & Unseen patterns \\
\hline
IRNet & 63.5 & 48.8 \\
IRNet $+$ Aug & 64.7 ($+1.2$) & 53.7 ($+4.9$) \\
\hline
RATSQL & 66.6 & 52.3 \\
RATSQL $+$ Aug & 73.0 ($+6.4$) & 55.4 ($+3.1$) \\
\hline
\end{tabularx}
\caption{EM accuracy over seen and unseen patterns on Spider.}
\label{tab:pattern_seen}
\end{table}

\subsection{Analysis}
\label{ssec:analysis}

\textbf{Case study}. To intuitively understand the advantages of HIER over FLAT, we present two typical examples in Table~\ref{tab:case study}. The FLAT approach fails to understand the column name ``loser\_age'' in the WHERE clause of the first SQL query, and overlooks the WHERE clause completely in the second query. In contrast, our HIER approach basically captures semantics of both two SQL queries, though the generated questions seem a little bit unnatural due to the ordering issue. Under our hierarchical generation approach, clause-to-subquestion translation is much simpler than direct SQL-to-question translation, hence leading to relatively high-quality NL questions.

\textbf{Component-level analysis}. To understand fine-grained impact of our proposed augmentation framework, we report CM F1 scores over five types of SQL clauses in Table~\ref{tab:complex_exp}. We observe that the main advantage of data augmentation on WikiSQL comes from the prediction of WHERE clause, which is also the main challenge of simple datasets. The performance of SELECT clause is near the upper bound, where most of the evaluation errors are due to wrong annotations by humans \cite{hwang2019comprehensive}.
For Spider, performances of all clauses are improved. Looking into the Spider dataset, we find that our generated subquestions are of high quality in the terms of diversity and semantics, e.g., ``age'' translated as ``from youngest to oldest'', and ``year'' as ``recent''.
It is interesting to see that performances of the right-side three types of complex clauses are much higher on DuSQL than on Spider, and also much higher than that of the basic SELECT/WHERE clauses on DuSQL itself. As discussed earlier in Section $\S$\ref{ssec:main_result}, we suspect this is because the complex clauses on DuSQL are more regularly distributed and thus more predictable due to their data construction method. 


\textbf{Analysis on SQL patterns}. One potential advantage of ASTG-based SQL generation is the ability to generate new SQL patterns that do not appear in the training data. To verify this, we adopt the more complex Spider, since its evaluation data contains a lot (20\%) of low-frequency SQL patterns unseen in the training data. We divide the question/SQL pairs into two categories according to the corresponding SQL pattern, and report EM accuracy in Table~\ref{tab:pattern_seen}. It is clear that our augmentation approach gains improvement both on seen and unseen patterns. The gains on unseen patterns show that with generated data as extra training data, the model possesses better generalization ability.

\textbf{Impact of augmented data size}. We study how the number of augmented pairs affects the accuracy of parsing models. We conduct this experiment on the Spider dataset using IRNet model based on the directly merging training strategy. In the experiment, we randomly sample question/SQL pairs from all the generated data based on multiples of the size of the original training data. Results are given in Table~\ref{tab:impact_size}. It is not surprising that more augmented data brings higher accuracy which is consistent with the observations in \citet{guo2018question}. Interestingly, we find that more augmented data brings more stable benefits.

\begin{table}[tb]
\small
\center
\begin{tabularx}{0.42\textwidth}{l|c|c|c|c}
\hline Size & 100\% & 200\% & 300\% & all\\
\hline
Acc & 59.1 & 59.4 & 59.3 & 61.8\\
 & ($\pm$1.18) & ($\pm$0.75) & ($\pm$0.69) & ($\pm$0.26)\\
\hline
\end{tabularx}
\caption{The impact of augmented data size on Spider using IRNet model. The numbers in brackets represent the variance of three runs.}
\label{tab:impact_size}
\end{table}

\begin{table}[tb]
\small
\center
\begin{tabularx}{0.48\textwidth}{l|r|r}
\hline Strategies & EM Accuracy & Total Training Time \\
\hline
Baseline & 59.5 & 6.9 hours \\
Pre-training & 60.0 & 36.1 hours \\
Directly merging & 61.8 & 34.9 hours \\
Sampling & 61.7 & 10.4 hours \\
\hline
\end{tabularx}
\caption{Comparison of three training strategies on Spider using IRNet model. The size of augmented data is about 6.7 times that of the original training data, as shown in Table ~\ref{tab:num_data}.}
\label{tab:train_method}
\end{table}

\textbf{Comparison on training strategies}. 
Table \ref{tab:train_method} compares the three training strategies for utilizing generated data, which are discussed in Section $\S$\ref{ssec:train_aug}. All experiments are run on one V100 GPU card. The pre-training strategy only slightly improves performance over the baseline, indicating that it fails to make full use of the generated data. The directly merging strategy and the sampling strategy achieve nearly the same large improvement. However, the sampling strategy is much more efficient.

\section{Related Work}
\label{sec:related-work}

\textbf{Data augmentation for NLP}. As an effective way to address the sparseness of labeled data, data augmentation has been widely and successfully adopted in the computer vision field \cite{szegedy2015going}. Similarly in the NLP field, a wide range of tasks employ data augmentation to accommodate the capability and need of deep learning models in consuming big data, e.g., text-classification \cite{wei2019eda}, low-resource dependency parsing \cite{csahin2018data}, machine translation \cite{fadaee2017data}, etc. Concretely, the first kind of typical techniques tries to generate new data by manipulating the original instance via word/phrase replacement \cite{wang2015s,jia-liang-2016-data}, random deletion \cite{wei2019eda}, or position swap \cite{csahin2018data,fadaee2017data}. The second kind creates completely new instances via generative models~\cite{yoo2019data}, while the third kind uses heuristic patterns to construct new instances \cite{yu2018syntaxsqlnet}.

\textbf{Data augmentation for semantic parsing}. Given an NL question and a knowledge base, semantic parsing aims to generate a semantically equivalent formal representation, such as SQL query, logic form (LF), or task-oriented dialogue slots.
Based on LF-based representation, \citeauthor{jia-liang-2016-data}~\shortcite{jia-liang-2016-data} train a synchronous context free grammar model on labeled data for generating new question/LF pairs simultaneously.
\citeauthor{hou2018sequence}~\shortcite{hou2018sequence} focus on the slot filling task. They train a Seq2Seq model on semantically similar utterance pairs, and generate new and diverse utterances for each original one. 

\textbf{Data augmentation for text-to-SQL parsing}. \citeauthor{iyer2017learning}~\shortcite{iyer2017learning} focus on in-domain text-to-SQL parsing. They automatically translate NL questions into SQL queries, and ask human experts to correct unreliable SQL queries. 
On Spider, ~\citeauthor{yu2018syntaxsqlnet}~\shortcite{yu2018syntaxsqlnet} collect many high-frequency SQL patterns and also convert corresponding questions into patterns by removing the concrete database-related tokens. They keep 50 high-quality question/SQL pattern pairs via manual check, and use them to generate new question/SQL pairs for a given table.\footnote{More specifically, given a pair of question and SQL query, they first manually align question tokens and DB elements, and replace the aligned terms with some special, generic symbols, resulting in question/SQL templates. Then given a new table, they generate question/SQL pairs by filling question/SQL templates with table elements.} However, their approach only considers SQL patterns concerning single table, and the need for human intervention seems expensive. 
~\citeauthor{guo2018question}~\shortcite{guo2018question} use a pattern-based approach to generate SQL queries and utilize a copy-based Seq2Seq model to directly translate SQL queries into NL questions. 
In contrast, this work proposes to use an ASTG for better SQL query generation and a hierarchical SQL-to-question generation approach to obtain higher-quality NL questions.

\section{Conclusions}

This paper presents a simple yet effective automatic data augmentation framework for cross-domain text-to-SQL parsing. With two-step processing, i.e., ASTG-based SQL query generation and hierarchical SQL-to-question generation, our framework is able to produce 
high-quality question/SQL pairs on given DBs. Results on three widely used datasets, i.e., WikiSQL, Spider, and DuSQL show that: 1) the hierarchical generation component is the key for performance boost, due to the more reliable clause-to-subquestion translation, and in contrast, previously proposed direct SQL-to-question generation leads to performance drop over strong baselines; 2) our proposed framework can consistently boost performance on different types of SQL clauses and patterns; 3) the sampling strategy is superior to the other two strategies for training parsers with both labeled and generated data, especially in the terms of training efficiency. 

\section*{Acknowledgments}

{\color{black}{We are very grateful to our anonymous reviewers for their helpful feedback on this work. Kun Wu, Zhenghua Li and Min Zhang were supported by National Natural Science Foundation of China (Grant No. 62176173 and 61876116), and a Project Funded by the Priority Academic Program Development (PAPD) of Jiangsu Higher Education Institutions.}}

\bibliography{emnlp2021}
\bibliographystyle{acl_natbib}

\end{document}